# A Consolidated Approach to Convolutional Neural Networks and the Kolmogorov Complexity

D Yoan Loic Mekontchou Yomba

**Abstract**

The ability to precisely quantify similarity between various entities has been a fundamental complication in various problem spaces specifically in the classification of cellular images. Contemporary similarity measures applied in the domain of image processing proposed by the scientific community are mainly pursued in supervised settings. In this work, we will explore the innovative algorithmic normalized compression distance metric (NCD) based on the information theoretic concept of Kolmogorov Complexity. Additionally we will observe its possible implementation in Convolutional Neural Networks (CNN) to facilitate and automate the classification of Retinal Pigment Epithelial cell cultures for use in Age Related Macular Degeneration Stem Cell therapy in an unsupervised setting. This metric has been successfully applied in computer graphics, vision, and image analysis but no previous work has pursued its possible applications in conventional unsupervised cognitive machines. This paper documents a practical study of this distance metric and aims to answer the following questions: Is there a relationship between the distance of image features in vector space and a learning machine's classification accuracy? Does the augmentation of the architecture of a CNN and the fusion of the normalized compression distance with a convolutional layer to elevate feature discrimination prove to be a formidable endeavor?

**Key Terms –** Kolmogorov Complexity, Statistical Pattern Recognition, Convolutional Neural Networks, Information Theory, Normalized Compression Distance, Image Similarity, Unsupervised Learning, Age Related Macular Degeneration

## I. Introduction

Age related Macular Degeneration (AMD) is a chronic progressive disease attributed to permanent loss of vision affecting over 100 million Americans. AMD is characterized by the formation of drusen at the central point of the macula stemming from the breakdown of retinal pigment epithelial (RPE) cells or the formation of blood vessels. A cure to this degenerative process has been persistently pursued by academics yielding much promise specifically through stem cell therapy. Stem cell therapy in the case of AMD aims to replace lost RPE cells in the retina through transplantation of progeny produced from stem cells. This replacement therapy requires precise analysis of the degree of cuboidal cobblestone morphology formation in order to identify cells that hold characteristics common to RPE. This characterization step is usually performed manually thereby introducing numerous sources of errors in the classification step. Thus, there is an immense need for an automated approach to the classification of progeny.

Previous research endeavors have employed several methods including discrete wavelets transforms [7], NCD computation over whole images [8], transfer learning in deep convolutional neural networks [10], as well as multi-instance learning with application to optical coherence tomography images [9], to tackle this research problem. While all approaches have achieved great classification accuracies, none have explored the integration of the NCD in an unsupervised learning space. Therefore, in this paper, we propose a new learning framework to facilitate the classification scheme of RPE cell cultures based on the observance of feature dissimilarity in vector space obtained through the precise application of the NCD. We test whether distinctly dissimilar features obtained from the CNN can better portray the class to which an image belongs. Additionally, we aim to observe any intuition a graphical representation of a distance matrix can portray in a lower dimensional space through spectral clustering.

## II. Outline

This paper is structured in the following manner, Section III provides a thorough review of the theory that validates the existence, applicability, and effectiveness of the normalized compression distance. Section IV explores the theory attributed to conventional CNNs. Section V proposes a novel method to adequately merge the effectiveness of the distance metric and the power of the convolutional layer. Section VI details our experimental methodology, results, and performance evaluation for the presented method. Lastly, Section VII presents conclusive remarks obtained throughout the endeavor.

## III. Kolmogorov Complexity Theory

The Kolmogorov complexity of an entity is often times described as the shortest length binary algorithm that entirely reproduces it in some descriptive form as output consequently defining the lowest magnitude of information needed from which the original entity could be completely retrieved from. Thus, given some input, its minimal description is figuratively modelled in the following manner,

$$K(input) = |D_{min}(input)|$$

Where $D_{min}$ symbolizes the minimal description length of the input.

From this, one can see that the reproducibility of an input from a lower magnitude representation is completely dependent on the degree of randomness exhibited by the patterns in the present input. Therefore, the Kolmogorov complexity aims to answer the question of, what is the level of regularity in the input bit patterns and how can we associate the most minimal description capturing the most pertinent characteristics of this? By doing so, it provides a true lower bound to the magnitude most compressors are able to achieve.

The Kolmogorov complexity can also be adapted to exploit a conditional compressive behavior in that the shortest length program to compute some entity can be conditioned on some supplementary input. Much work has been done in this domain specifically by Vitanyi [1] who concisely showed the dependence of the Kolmogorov complexity on the coding theorem and derived a beautiful proof connecting the conditional coding theorem and universal semi probability mass functions in the following fashion

$$K(input|auxiliary\ input) = \log\left(\frac{1}{m(input\ |auxiliary)}\right) = \log\left(\frac{1}{Q_u(input\ |auxiliary)}\right)$$

Where m and $Q_U$ serves as a semi probability mass functions to different reference machines.

This descriptive complexity led to further work on compression based similarities by Bennett et al.[2] in which the shortest length program needed to transform one string to another dissimilar one quantified the two strings' information distance. Li et al. [3] expanded on this work and provided a universal normalized version of the information distance as the similarity metric. Which in turn led to the formulation of the Normalized Compression Distance by Li et al. [4]

$$NCD(x, y) = \frac{C(xy) - \min\{C(x), C(y)\}}{\max\{C(x), C(y))\}}$$

Where C(x) delimits a compressive scheme on some input x and XY defines the concatenation of two objects. This metric satisfies the triangle inequality, positivity argument, and is symmetrical. Vitanyi and Cilibrasi [5] expanded on this work to provide bounds on this "normal compressor" in which the NCD is bounded by $[0, 1 + e]$. 0 is definitive of the highest order of similarity between two entities while $1 + e$ defines maximal dissimilarity. The additive error models compressor imperfections in modern day applications.

This distance metric has achieved great success particularly in algorithmic music clustering [6] and genomic sequencing [5].

## IV. Convolutional Neural Networks

Convolutional neural networks are beautiful complex learning machines and have achieved a great deal of success on challenging visual classification tasks. They are comprised of a multitude of neurons organized as multilayer perceptrons and aim to replicate the visual cortex. Individual neurons within such architectures lack the interconnectivity to past layer activations of typical neural networks and are associated to specific receptive fields of an input thus, each neuron only responds to stimuli stemming from their region of local connectivity. This niche architecture negates the need of feature space priori knowledge making such systems extremely advantageous especially for unsupervised learning tasks.

CNNs are normally comprised of a multitude of repeating units in a layered representation performing different learning tasks in the most optimal fashion. Such layered components include convolutional, softmax, pooling, dropout, and fully connected layers as well as various activation functions. Figure 1 depicts an adequate demonstration of the anatomy of this system.

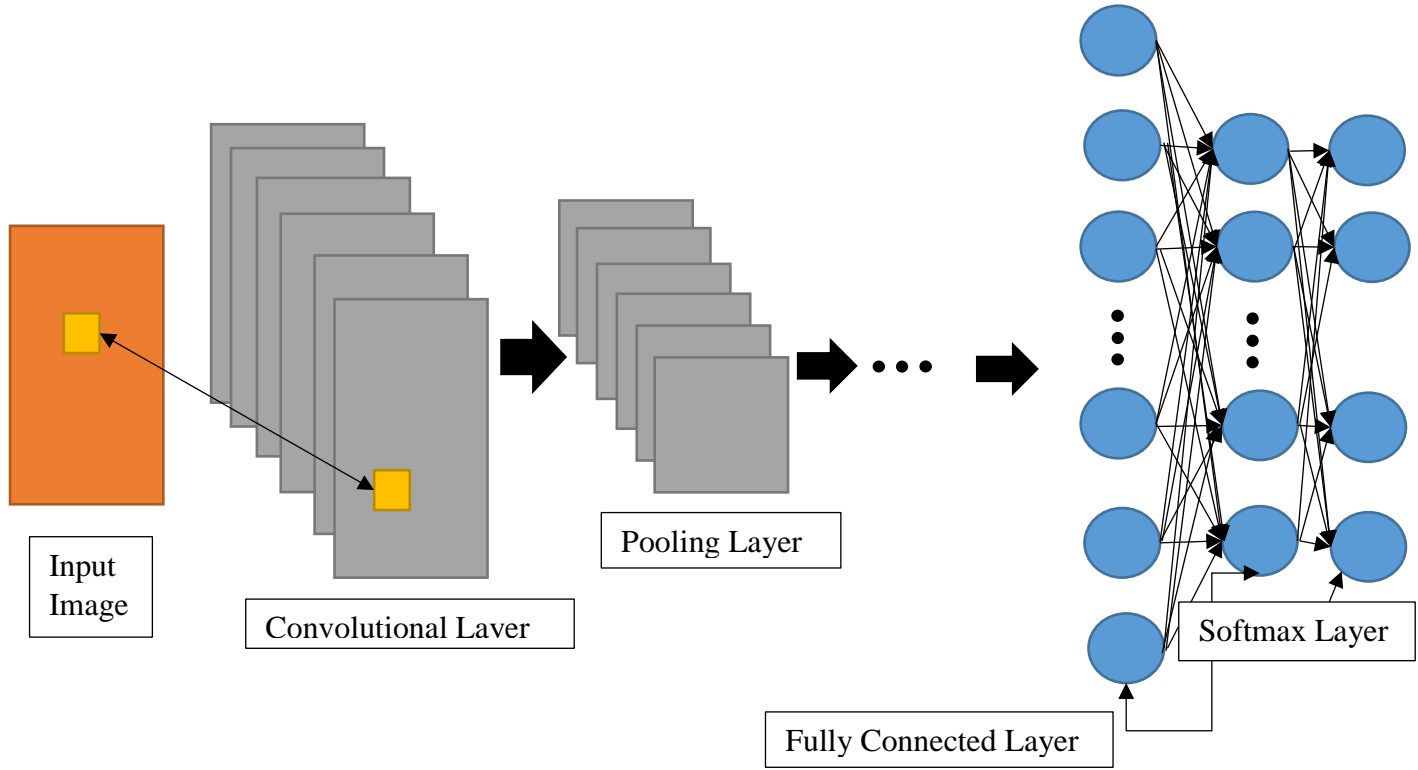

*Figure 1: Anatomy of a CNN*

Convolutional layers mirror the response of a neuron to some stimuli originating from a receptive field. The layers are comprised of an input $I$, filters $K$ of size $K_1 * K_2$ denoting the size of the receptive field, a stride $s_i$ stating the distance between two receptive fields, and a bias term $b_i$. The input to a convolutional layer is normally a three dimensional object. For example in the case of images, the first two dimensions may encompass the height $H$ and width $W$ of an image while the third denotes the number of color channels $C$ which is typically three. Such inputs can be expressed in a three dimensional space $R^{H*W*C}$. The output after such a convolutional operation is applied is modelled below.

$$(I * K)_{ij}^{output} = b_i + \sum_{m=0}^{K_1-1} \sum_{n=0}^{K_2-1} \sum_{C=1}^{C} K_{m,n,c} * I_{i+m,j+n,c}$$

It is important to note that normal convolutional layers have a multitude of filters which creates a filter representation inherently lying in a 4th dimensional space $R^{K_2*K_1*C*|K|}$ which cohesively forms a 4th order feature map as output.

The pooling layer is mainly interested in performing some down sampling operation on an input and constraining its size. Such layers typically rely on a receptive field of some initially defined size and replace each region by a singular value computed by some operation to achieve the preferred spatial reduction. Present below are typical operations performed by such layers.

$$\frac{\sum_{i=1}^{n} I_i}{n}, \frac{\partial g}{\partial I} = \frac{1}{m} \dashrightarrow Mean\ Pooling$$

$$\max(x), \frac{\partial g}{\partial I} = \begin{cases} 1 * I \text{ if } I_i = \max(I) \\ 0 * I \text{ otherwise} \end{cases} \dashrightarrow \text{Mean Pooling}$$

n denotes the size of the pooling layer, $I$ denotes the input, y the output, and g denotes a function/operation which in this case is a filter.

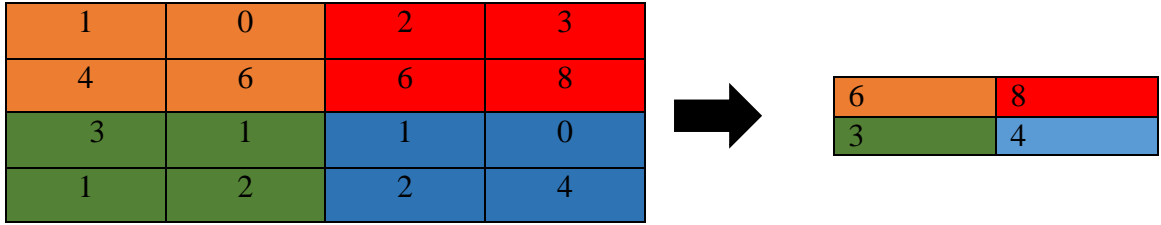

*Figure 2: example of Max Pool Operation*

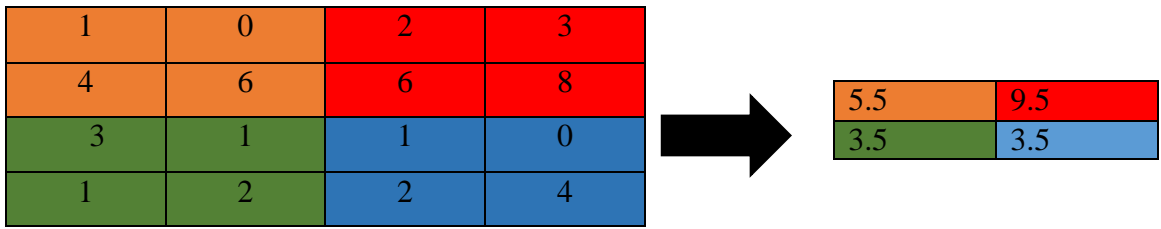

*Figure 3: example of Mean Pool Operation*

The dropout layer is useful when training specifically to avoid overfitting a model to the training set. Half of the neurons at a specific layer are deactivated during training in order to elevate generalization forcing neurons to adapt and learn the same basic interpretation of some input. The dropout layer is then deactivated during the phase of prediction.

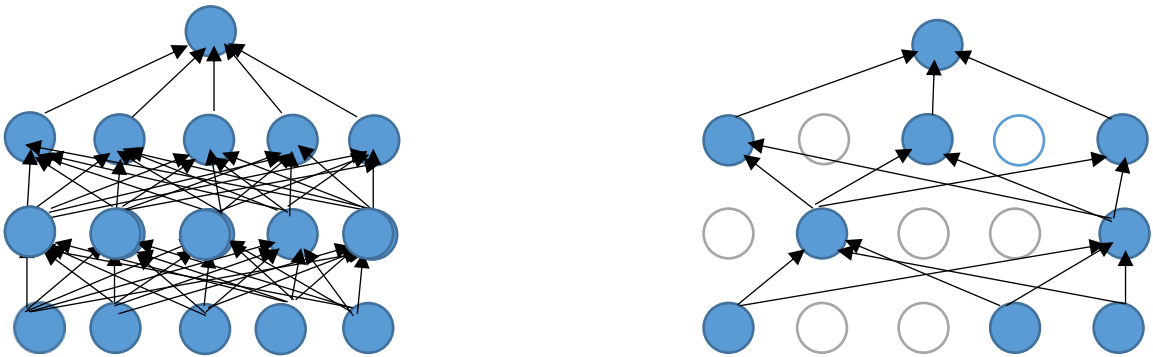

Figure 4. Hidden Layers without Dropout and with Dropout

Fully connected layers are similar to hidden layers in conventional neural networks as neurons are connected to all past activations present in a previous layer. The scheme is useful in calculating its activations, properly classifying an input, and performing forward and backward propagation.

Lastly, the softmax layer serves as the output layer of the CNN. It computes a probability vector denoting the probability distribution of the possible classes to which an input may lie in. The maximal value in this probabilistic vector is observed and mapped to a real value denoted as the class of the input.

CNNs have had successfully applications in image classification [12], facial verification [13], and audio classification [11].

## V.    Novel Approach

Convolutional neural networks although very powerful hold pertinent drawbacks. Sample drawbacks include the need for a high order of training data, high computational costs, and the loss of features at the pooling layer. We propose a new convolutional layer with the goal of propagating only the most independent and information rich feature vectors for classification and significantly reducing the overall computational overhead present in most architectures by relinquishing the need for the pooling layer.

Normally, the pooling layer performs some operation on an input feature map to reduce its spatial complexity thereby constricting a subspace's spatial representation by a computing singular value for a given receptive field. These means of computation usually apply the maximal value or the average of all values in an input's subspace which poses the risk of either selecting a numerical value that doesn't adequately represent the receptive field or losing out on various other important spatial representations crucial to accurate prediction at the output layer. Such loss of information or misrepresentation of information is alarming and we believe the use of the NCD at the convolutional layer will not only resolve this issue but obtain and propagate only the richest feature vectors.

In order to adequately utilize the NCD, the formulation of an output feature map comprised of the most independent feature vectors is performed in two steps: the convolutional step and the vector space distance computation step.

In the convolutional step, normal convolution is applied to an input with the result having a spatial representation explained by the following equation.

$$Spatial\ output\ size = \frac{(W - F + 2 * P)}{S} + 1$$

W denotes the size of the input feature map, F the size of the filter, P the magnitude of the padding and S the magnitude of the stride.

In the computation of the distance in vector space, the BZIP algorithm is selected to compress the feature vectors. The NCD is then computed over such feature vectors. However because we are dealing with objects in feature space, the NCD equation must be manipulated and is done below as proposed by Sculley et al.[13].

$$NCD(X,Y) = \frac{C(XY) - \min(C(X), C(Y))}{\max(C(X), C(Y))} \rightarrow Transformation\ in\ Vector\ Space$$

$$\rightarrow 1 - \left(1 - \frac{C(XY) - \min(C(X), C(Y))}{\max(C(X), C(Y))}\right)$$

$$\rightarrow 1 - \left(\frac{\max(C(X), C(Y))}{\max(C(X), C(Y))} - \frac{C(XY) - \min(C(X), C(Y))}{\max(C(X), C(Y))}\right)$$

$$\rightarrow 1 - \frac{C(X) + C(Y) - C(XY)}{\max(C(X), C(Y))} \rightarrow 1 - \frac{|X| + |Y| + |X+Y|}{|X+Y|}$$

X and Y are input Feature vectors preferably in uint8 byte array form for ease of computation.

The distance matrix is then constructed from the application of the NCD and represented in the following manner,

$$distance\ matrix = \begin{pmatrix} x_{1,1} & \cdots & x_{1,n} \\ \vdots & \ddots & \vdots \\ 0 & \cdots & x_{n-1,n} \end{pmatrix}$$

$x_{i,j}$ represents the compression distance between vector i and j. Note the diagonal of the distance matrix will be comprised of 0 as the Kolmogorov complexity of an entity to itself is 0.

From the distance matrix, feature vectors holding a distance higher than some pre-defined dissimilarity criterion can be used to form a feature map. If no distance between any two feature vectors satisfies the dissimilarity argument, we propagate the feature vectors with distances greater than the mean of all distances. Algorithm 1, present below concisely presents these methods and steps.

---

**Algorithm 1** Convolutional Layer

---

**Notation:** $K_{i,j}$ is representative of the number of weight/filters at a given convolutional layer of size $K_1 * K_2$ denoting the size of the receptive field, a stride $s_i$ states the distance between two receptive fields, $F(x)$ represents an activation function which in this case is the sigmoid function $\frac{1}{1+e^{-Bx}}$, and a bias term is expressed as $b_i^{Layer}$. The input to a convolutional layer is normally a three dimensional object $I_i^{Layer}$ or a four dimensional feature map represented by $I_{j,m}^{Layer}$. $I_{j,m}^{Layer}$ models a feature map however, at each iterative step of convolution, it serves as the $m^{th}$ unit of the $j^{th}$ feature map we are convolving. Lastly, $F$ is representative of the size of the filter..i.e. $F = K_1 * K_2$.

**Layer Parameter Configurations:** Stride, Filter Size, # of Filters, Padding, and Dissimilarity Criterion

**Input:** Input Image of size $H \times W \times C$ or Input Feature Map of size $H \times W \times C \times K$ where $K$ signifies the magnitude of the feature map's 4$^{th}$ dimension

**Output:** Output Feature Map

1. Perform Convolution
   a. Input Image
      i. $I_i^{Layer+1} = f(b_0^{Layer} + \sum_{m=0}^{F-1} \sum_{C=1}^{C} K_{m,c} * I_{i,m}^{layer}$, i in this case signifies the number of input images which in this case is 1.
   b. Input Feature Map
      i. $I_{j,m}^{Layer+1} = \sum_{i=1}^{|K|} (f(\sum_{n=0}^{F-1} \sum_{C=1}^{C} K_{i,j,n,c} * I_{i,m+n-1}^{layer} + b_{0,j}^{Layer})$ as proposed by Osama et al.[15];
   c. $|I_i^{Layer+1}$ or $I_{j,m}^{Layer+1}| = \frac{(W-F+2*P)}{S} + 1$
2. Perform Compression in Vector Space
   a. **FeatureMat = Compress($I_i^{Layer+1}$, BZIP)** //compressing each feature vector and storing in a matrix
3. Compute NCD distance matrix on Individual Feature Vectors
   a. **for** i = 1:number of columns in FeatureMat **do**
   b.    **for** j = i+1: number of columns in FeatureMat **do**
            Vector1 = FeatureMat (:,i); // *acquiring the i$^{th}$ feature vector*
            Vector2 = FeatureMat (:,j); // *acquiring the j$^{th}$ feature vector*
            distance matrix(i,j) = NCD(Vector1, Vector2) // *compute the NCD*

   c. Distmean = **DistanceMatrixMean**(distance matrix) // compute the mean value of the distance matrix from all cells not lying on diagonal
   d. **for** i=1:length(distance matrix(rows)) **do**
   e.    **for** j = i+1:length(distance matrix(rows)) **do**
   f.       **if** (distance matrix(i,j) > criterion)
   g.          output feature map = $[I_i^{Layer+1}(:,i), I_i^{Layer+1}(:,j)]$ // *generate feature map*
   h. output feature map = unique(output feature map) // *keep only the unique feature vectors*
   i. **if**( output feature map is empty)
   j.    criterion = Distmean; // *assign mean distances as dissimilarity criterion*
   k. **Repeat:** step d – h;
4. Propagate the feature map
   a. $I_{j,m}^{Layer+1}$ = output feature map

---

The vector space distance computation step holds a computational complexity of $O(N^2)$. While the convolutional step exhibits a computational complexity of $O(WHkk)$. W and H denote the height and width of an image and $(k_1, k_2)$ depict the size of the receptive field or filter. The complexity of convolution could be optimized to $O(2 * WHk)$ if separable convolution is applied as proposed by [16]. Thus the worst case complexity of this approach is $O(N^2) + O(WHkk)$ with H, W, and k holding constant values.

It is important to note that standard forward, backward pass, and parameter update steps will be unaffected by the introduction of the above layer.

## VI. Experiments

The experimental framework comprised of image processing, training set duplication, CNN internal parameter tuning, NCD feature computation, and support vector machine validation steps. In the image processing step, we applied denoising algorithms and filters to the images to remove unneeded noise and converted colored 3-channel input images to their black and white 1-channel complements. The training set duplication step entailed duplicating the training set as our initial image set comprised of only 34 images stemming from 5 classes which are present below in figures 5 - 8. To generate more training data from the initial sample set, various methods were applied such as image complements, image partitioning, translational, as well as rotational techniques.

In the parameter tuning step, the internal parameters of the CNN were tuned by maintaining a Gaussian process model to minimize an objective function which in this case was the classification error rate on the validation set through the utilization of Bayesian optimization. The CNN's depth, initial learning rate, and momentum were variables chosen for optimization. The CNN's architecture was configured to maintain uniform computation at each layer. To accomplish this endeavor, padding was added to all convolutional layers to maintain similar input and output sizes. The number of filters were also chosen proportionally based on the depth of the network so that the magnitude of parameters remained uniform throughout various objective evaluations. Additionally, data augmentation techniques were administered during training to prevent from overfitting. Thirty objective evaluations were performed resulting in a deep architecture comprised of 63 layers, a learning rate of 0.00048871, and a momentum of 0.907. The generated CNN was used on a test set of 15 images yielding a 45.54% classification error rate. Such a classification error rate is fairly high but is completely dependent on the magnitude of training images given to the CNN as only 232 images stemming from 5 classes were used in our the training step.

A feature map was then extracted from the middle most and last convolutional layer originating from the hypothesis that concise representations of the input image would have been realized at this period in the architecture. The NCD was then applied inherently generating a distance matrix. From this distance matrix, a dissimilarity criterion of 0.4 was selected to observe similarity based on the most dominant shared features in vector space. Feature vectors with distances of magnitudes greater than the dissimilarity criterion were thus obtained for further processing. The last convolutional layer held maximally similar feature vectors thus, no features were selected so the middle most convolutional features were used instead. The distances of 232 feature vectors were initially computed with 43 vectors holding maximal dissimilarity from one another.

Two identical support vector machines were then generated. The 232 initial feature vectors were used to train the first support vector machine while the 43 features obtained from the NCD computation were used to train the second support vector machine. A validation set of 15 images was then applied to both machines. Both support vector machine held a classification error rate of 45.54%. Although such an error rate is fairly high and completely dependent on the features obtained from the middle most convolutional layer, it provides us with very pertinent findings. We are able to see that the 43 independent feature vectors obtained from the application of the

NCD hold the most pertinent information specific to the class to which an image may belong to. Thus, further validating the differential ability of the NCD in vector space and its potential use in conventional CNN.

The information present within the distance matrix was also of high interest to us. We aimed to observe any intuition specific to the number and separability of classes present within the distance matrix in a graphical space. To accomplish such an endeavor, it was deduced that spectral clustering would be an interesting application to observe such underlying behaviors.

The CNN obtained from the Gaussian process model was duplicated and utilized to train one image from each class. Features were then extracted from each CNN and the NCD was applied to all 5 feature vectors to generate a distance matrix currently present in figure 9. From the distance matrix, the affinity matrix was constructed to observe the connectivity of the distance matrix in graphical space. The degree matrix was then computed by summing over each row of the affinity matrix. By taking the difference of the degree and affinity matrix, the graph laplacian was constructed. The graph laplacian is typically useful to observe various properties of a graph including the number of spanning trees present.

Eigen decomposition was then performed on the graph laplacian to reduce its dimensionality resulting in a set of eigenvalues and eigenvectors. The eigenvalues were then sorted in ascending order and the second smallest eigenvalue was selected with hopes of observing the algebraic connectivity of the graph in order to approximate its sparsest cut.

Spectral clustering was then performed by making use of the 2nd and $3^{rd}$ smallest eigenvalues and their constituent eigenvectors to portray behaviors in a lower dimensional space. This lower dimensional representation of the graph laplacian is present below in figure 13-14. From further observation, it is easy to discern that the $2^{nd}$ and $3^{rd}$ smallest eigenvalues concisely demonstrate the connectivity of the distance matrix as well as adequately partitioning the graph in 5 places. Thus, we can infer that the distance matrix withholds various underlying behaviors including the number of classes currently represented by the distances lying within it.

## VI.     Conclusive Remarks

In this paper we present a novel approach aimed at fusing the effectiveness of the NCD and the power of the convolutional layer. We devise an algorithmic scheme and observe its potential worst case complexity. We observe that by adequately harnessing the precision of the NCD and employing this distance metric in vector space, we can effectively better preserve the features withholding the richest information for prediction. We conclude that the augmentation of the architecture of a CNN and the fusion of the normalized compression distance with a convolutional layer to elevate feature discrimination is a formidable endeavor. Additionally, we conclude that the computational overhead in training a cognitive agent from the feature vectors acquired through the NCD is much lower than the case in which features are not processed by the NCD. From a lower magnitude of information, we are able to achieve equal classification error rates as machines with higher orders of data. We however cannot accurately quantify the relationship between the distance of image features in vector space and a learning machine's classification accuracy thus forming grounds for further work to be performed in the future. Moreover, in the future, we will observe 'smart' methods to select the most optimal dissimilarity metrics as well as explore a compression scheme based on the estimation of the feature vector's underlying probability distribution for elevated NCD precision.

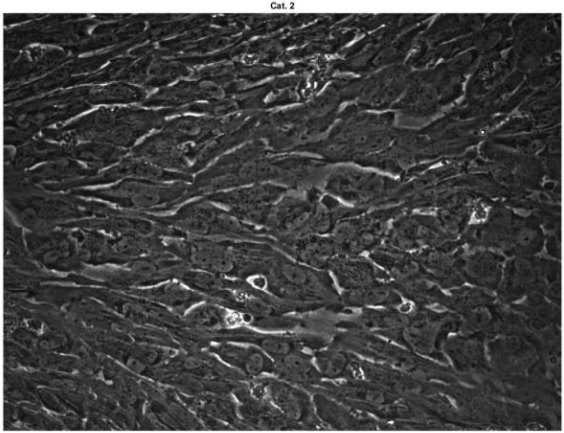

Figure 5: RPE Cell Cultures - Category 1

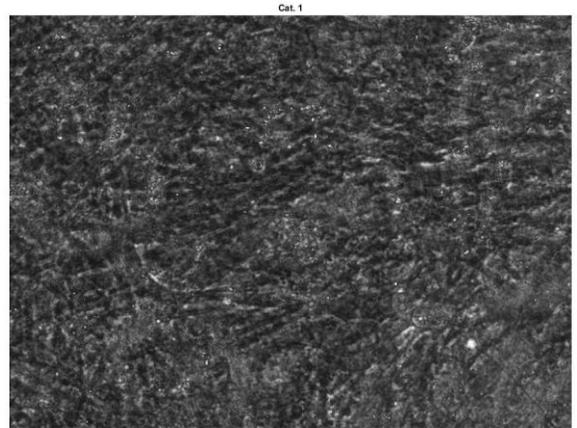

Figure 6: RPE Cell Cultures - Category 2

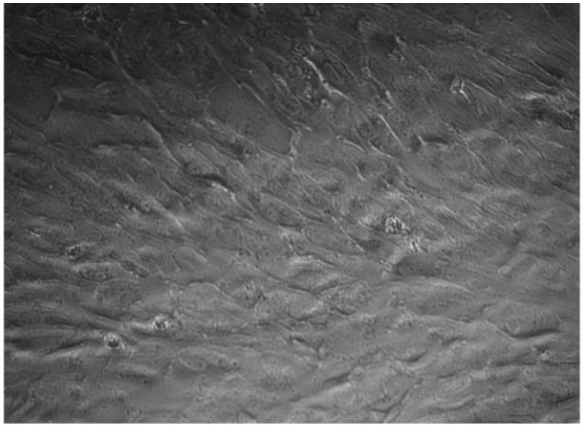

Figure 7: RPE Cell Cultures - Category 3

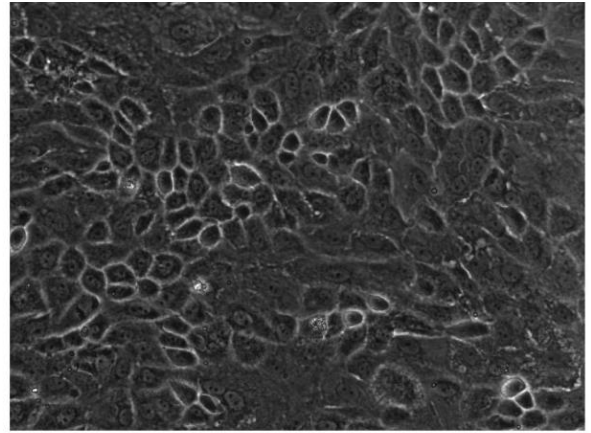

Figure 8: RPE Cell Cultures - Category 4

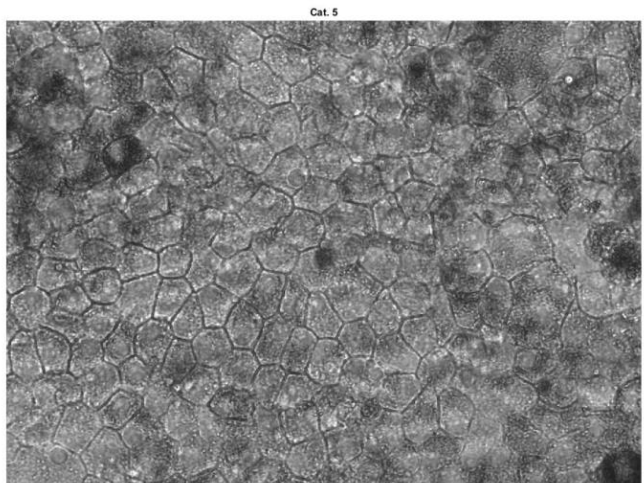

Figure 9: RPE Cell Cultures - Category 5

|   | A | B | C | D | E |
|---|---|---|---|---|---|
| A | 0.00000 | 0.49775 | 0.49787 | 0.49787 | 0.24891 |
| B | 0.49775 | 0.00000 | 0.49791 | 0.49782 | 0.24895 |
| C | 0.49787 | 0.49791 | 0.00000 | 0.49794 | 0.24895 |
| D | 0.49787 | 0.49782 | 0.49794 | 0.00000 | 0.24895 |
| E | 0.24891 | 0.24895 | 0.24895 | 0.24895 | 0.00000 |

*Figure 9: Distance Matric of Images Stemming From 5 Classes*

**degree matrix**

|   | A | B | C | D | E |
|---|---|---|---|---|---|
| A | 4.08397 | 0.00000 | 0.00000 | 0.00000 | 0.00000 |
| B | 0.00000 | 4.08397 | 0.00000 | 0.00000 | 0.00000 |
| C | 0.00000 | 0.00000 | 4.32714 | 0.00000 | 0.00000 |
| D | 0.00000 | 0.00000 | 0.00000 | 4.32720 | 0.00000 |
| E | 0.00000 | 0.00000 | 0.00000 | 0.00000 | 4.17870 |

*Figure 10: Degree Matrix*

**graph laplacian**

|   | A | B | C | D | E |
|---|---|---|---|---|---|
| A | 3.08397 | -0.70331 | -0.77963 | -0.77963 | -0.82141 |
| B | -0.70331 | 3.08397 | -0.77962 | -0.77965 | -0.82140 |
| C | -0.77963 | -0.77962 | 3.32714 | -0.99996 | -0.76794 |
| D | -0.77963 | -0.77965 | -0.99996 | 3.32720 | -0.76796 |
| E | -0.82141 | -0.82140 | -0.76794 | -0.76796 | 3.17870 |

*Figure 11: Graph Laplacian*

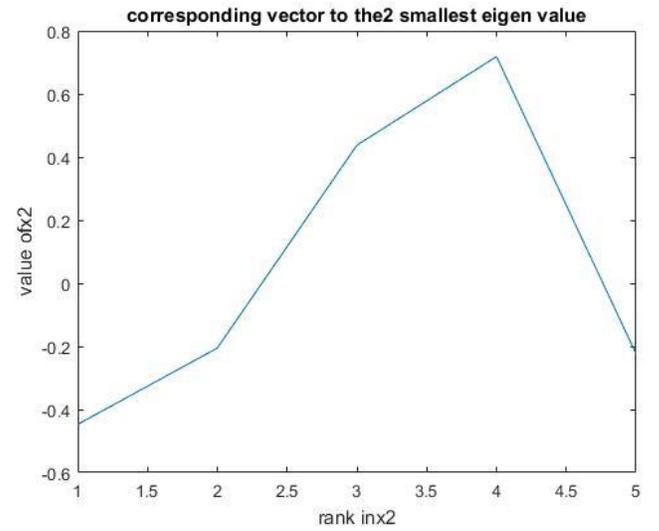

Figure 12: Eigen Vectors

Figure 13: Spectral Clustering based on the 2nd Eigen Value

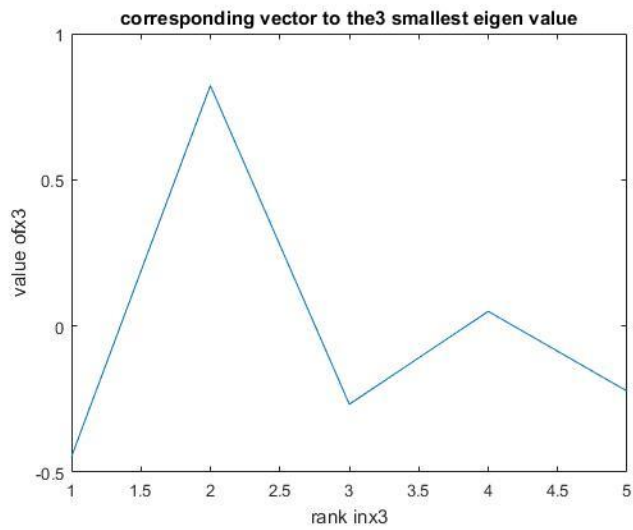

Figure 14: Spectral Clustering based on the 3rd Eigen Value